\documentclass{article} 
\usepackage[T1]{fontenc}
\let\articleaddcontentsline\addcontentsline
\usepackage{iclr2027_conference,times}
\let\addcontentsline\articleaddcontentsline

\usepackage{amsmath,amsfonts,bm}

\def\eqref#1{equation~\ref{#1}}

\def\1{\bm{1}}

\DeclareMathAlphabet{\mathsfit}{\encodingdefault}{\sfdefault}{m}{sl}
\SetMathAlphabet{\mathsfit}{bold}{\encodingdefault}{\sfdefault}{bx}{n}

\usepackage{amsmath}
\usepackage{amssymb}
\usepackage{booktabs}
\usepackage{multirow}
\usepackage{graphicx}
\usepackage{float}
\usepackage{afterpage}
\usepackage{hyperref}
\usepackage{url}
\usepackage{xspace}

\newif\ifappendixcontents
\let\linkedaddcontentsline\addcontentsline
\renewcommand{\addcontentsline}[3]{%
  \ifappendixcontents\linkedaddcontentsline{#1}{#2}{#3}\fi}

\title{RIGen: Beyond Action Following with Reciprocal Interaction in Generative Gameplay}

\author{\textbf{Mengfei Li$^{1,2}$}\thanks{Contact email: \texttt{mengfei.murphy.li@gmail.com}.}, \textbf{Yi Zhang$^{2}$}\thanks{Project Lead.}, \textbf{Xingtong Ge$^{1,2}$}, \textbf{Lunjie Zhu$^{1,2}$}, \textbf{Fangyu Lin$^{1,2}$}\\
\textbf{Peng Li$^{1}$}, \textbf{Guanglu Song$^{2}$}, \textbf{Yu Liu$^{2}$}, \textbf{Wei Xue$^{1}$}\thanks{Corresponding author.}, \textbf{Wenhan Luo$^{1}$}, and \textbf{Yike Guo$^{1}$}\\
\textnormal{\textsuperscript{1}The Hong Kong University of Science and Technology} \\
\textnormal{\textsuperscript{2}Vivix Group Limited}}

\newcommand{\rig}{RIGen\xspace}
\newcommand{\rigarena}{RIGen-Arena\xspace}

\iclrfinalcopy

\begin{document}
\pagestyle{plain}       
\thispagestyle{plain}   

\maketitle

\afterpage{\begin{table}[!t]
  \centering
  \caption{\textbf{Capabilities of generative gameplay systems.}
  Player$\to$World denotes player actions changing the state of objects or
  characters in the world. World$\to$Player denotes world events changing
  the player's gameplay state, such as health or available actions.
  NPC planning denotes explicit scene- and state-conditioned behavior reasoning.
  New scenes tests transfer to unseen scenes.}
  \label{tab:gameplay_capabilities}
  \vspace{2pt}
  \fontsize{8}{9.5}\selectfont
  \setlength{\tabcolsep}{1.3pt}
  \renewcommand{\arraystretch}{1.08}
  \newcommand{\capmethod}[2]{#1~{\fontsize{7}{8}\selectfont\citep{#2}}}
  \newcommand{\caphead}[2]{\parbox[c][18pt][c]{#1}{\centering #2}}
  \begin{tabular*}{\linewidth}{@{\extracolsep{\fill}}lccccccc@{}}
    \toprule
    & \multicolumn{2}{c}{\textbf{Control}}
    & \multicolumn{3}{c}{\textbf{Gameplay}} & \multicolumn{2}{c}{\textbf{Output}} \\
    \cmidrule(lr){2-3}\cmidrule(lr){4-6}\cmidrule(l){7-8}
    \textbf{Method} & \caphead{36pt}{Navigation} & \caphead{25pt}{Actions}
    & \caphead{49pt}{Player$\to$World} & \caphead{49pt}{World$\to$Player}
    & \caphead{29pt}{NPC\\Planning} & \caphead{20pt}{Audio}
    & \caphead{27pt}{New\\Scenes} \\
    \midrule
    \capmethod{GameNGen}{valevski2025gamengen}
      & $\checkmark$ & $\checkmark$ & $\checkmark$ & $\checkmark$ & -- & -- & -- \\
    \capmethod{SCOPE}{tong2026scope}
      & $\checkmark$ & $\checkmark$ & -- & -- & -- & -- & $\checkmark$ \\
    \capmethod{LingBot-World 2.0}{gao2026lingbot}
      & $\checkmark$ & $\checkmark$ & $\checkmark$ & -- & -- & -- & $\checkmark$ \\
    \capmethod{Incantation}{zhu2026incantation}
      & $\checkmark$ & $\checkmark$ & $\checkmark$ & $\checkmark$ & -- & -- & -- \\
    \capmethod{HelloWorld}{ouyang2026helloworld}
      & $\checkmark$ & $\checkmark$ & -- & -- & -- & $\checkmark$ & $\checkmark$ \\
    \capmethod{EchoWM}{zhang2026echowm}
      & $\checkmark$ & -- & -- & -- & -- & $\checkmark$ & $\checkmark$ \\
    \midrule
    \textbf{\rig (Ours)}
      & $\checkmark$ & $\checkmark$ & $\checkmark$ & $\checkmark$ & $\checkmark$ & $\checkmark$ & $\checkmark$ \\
    \bottomrule
  \end{tabular*}
\end{table}
  \afterpage{\begin{figure}[!t]
    \centering
    \includegraphics[width=\linewidth]{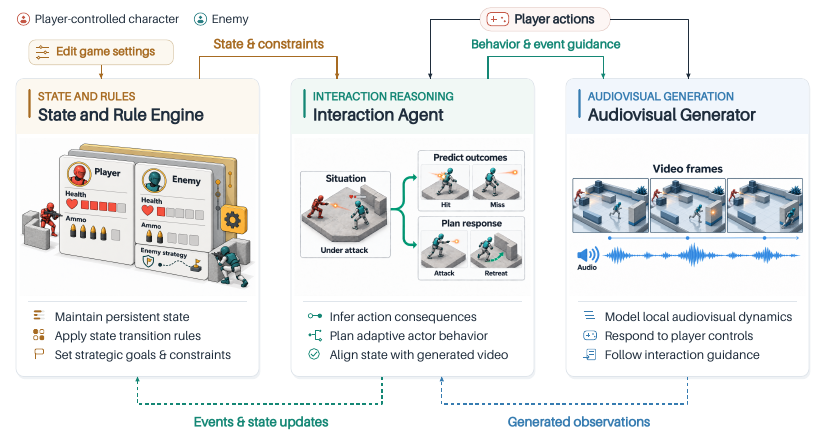}
    \caption{\textbf{Three coupled responsibilities in \rig.}
    State and rules maintain persistent gameplay state, apply transition rules,
    and expose editable game settings and strategic constraints. Interaction
    reasoning infers action consequences and plans behavior. Audiovisual
    generation realizes local dynamics under player controls and guidance,
    returning observations for subsequent decisions.}
    \label{fig:causal_hierarchy}
\vspace{-6pt}
\end{figure}
}}

\begin{figure}[H]
  \centering
  \includegraphics[width=0.90\linewidth]{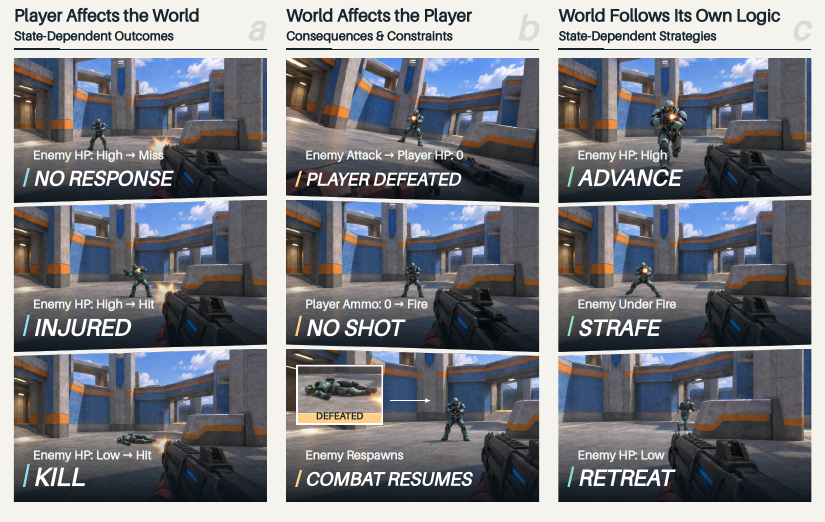}
  \caption{\textbf{Reciprocal interaction in \rig.}
  Players affect the generated world, whose rules, persistent state, and
  characters in turn shape gameplay.
  (a) Attacks have state-dependent consequences.
  (b) Enemy actions, resource constraints, and respawn rules shape the
  player's survival and continued play.
  (c) Opponents adapt their behavior to their health and current combat
  conditions.}
  \label{fig:teaser}
\end{figure}

\begin{abstract}
Generative gameplay requires more than immediate responses to player actions.
Actions should have persistent consequences, while the generated world should
act according to its own state and rules, shaping the player's situation and
subsequent choices. Video generation captures local dynamics, but limited
history, partially observed game state, and the need for contextual behavior
decisions make such interactions difficult to sustain through audiovisual
prediction alone. We introduce \rig (Reciprocally Interactive Generation), a framework for
reciprocal interaction in generative gameplay. \rig combines persistent state and configurable
rules, contextual interaction reasoning, and action-conditioned audiovisual
generation. Together, these components make interaction consequences and
actor behavior jointly determine generated content while maintaining
consistency between game state and visible outcomes. We construct a Xonotic
gameplay dataset to train an audiovisual generator for learning dynamics and
interactions. Further post-training enables \rigarena, an interactive generative
first-person shooter (FPS) prototype.
Experiments demonstrate more reliable combat outcomes and reciprocal
interaction between the player and the generated world.
Project page: \url{https://murphylmf.github.io/RIGen/}.

\end{abstract}

\section{Introduction}
\label{sec:introduction}

Generative models offer a route to game experiences whose environments,
characters, and events can be created during
play~\citep{bruce2024genie,yu2025gamefactory}.
Such experiences could be personalized to each player's preferences and
choices. Realizing this potential requires reciprocal interaction: players
should be able to change the world through their actions, while the world
follows its own operating logic. Its characters act according to their
situation and goals, and its rules and events shape the player's
circumstances and subsequent choices.

Recent video-based world models have made progress toward this goal.
DIAMOND~\citep{alonso2024diamond} and GameNGen~\citep{valevski2025gamengen}
demonstrate playable game simulation.
LingBot-World~\citep{robbyant2026lingbotworld} and SCOPE~\citep{tong2026scope}
provide responsive camera and movement control, while
LingBot-World~2.0~\citep{gao2026lingbot} and
Incantation~\citep{zhu2026incantation} incorporate semantic guidance for
character behavior and interaction events.
Table~\ref{tab:gameplay_capabilities} compares these systems across control,
gameplay, and output capabilities. It highlights the remaining gap in combining
reciprocal state changes with scene- and state-conditioned actor planning.

The prevailing approach builds on video models that capture rich local dynamics
and generate continuous audiovisual feedback conditioned on player controls,
making them a natural foundation for interactive games. However, sustaining reciprocal interaction
through audiovisual prediction alone faces three challenges.
\textbf{(1) Coordinating reciprocal interaction.} Responding to player actions
does not by itself determine how the world should act upon the player.
Actors must choose behavior based on current conditions and their own goals,
and the consequences must update player state and constrain subsequent
interactions.
\textbf{(2) Limited history.} Events beyond the model's context window can
still have consequences for current interactions~\citep{xiao2025worldmem}.
\textbf{(3) Partially observed state.} Gameplay variables such as health and
ammunition may not be recoverable from the visible scene, yet determine
whether an action can be executed and what outcome it should have.

In sustained combat, for example, accumulated damage must determine whether
the next hit is lethal, even when earlier attacks fall outside the video
context. The opponent's remaining health and goals should shape whether
it attacks or retreats. Damage dealt by the opponent must likewise affect
the player's health and ability to continue fighting.

To address these challenges, we introduce \rig, a framework for reciprocal
interaction in generative gameplay (Figure~\ref{fig:causal_hierarchy}).
The \emph{state and rules} layer maintains persistent gameplay variables and
configurable rules, including strategic goals and constraints.
The \emph{interaction reasoning} layer interprets the scene and player actions
to infer events and plan actor behavior under these conditions.
The \emph{audiovisual generation} layer realizes the resulting guidance
alongside player controls. Together, these layers let player actions change
the state that governs subsequent interactions. The world also acts
according to its own rules and goals, affecting the player's state and
available actions.

We instantiate \rig in first-person shooter (FPS) environments.
Our experiments assess combat outcomes, while qualitative demonstrations
and \rigarena illustrate state-dependent interactions and autonomous
enemy behavior that affects the player (Figure~\ref{fig:teaser}).

Our contributions are threefold.
\textbf{(1) A system for reciprocal generative gameplay.} We integrate
interaction prediction, executable state updates, and audiovisual generation
into a closed loop in which player actions change world state and world
behavior affects the player.
Model comparisons and controlled ablations demonstrate more reliable
realization of state-dependent interaction outcomes.
\textbf{(2) FPS gameplay dataset.} We construct a dataset covering
basic dynamics, controlled interactions, and sustained combat, pairing
audiovisual observations with player controls and interaction annotations,
with a reproducible data creation pipeline.
\textbf{(3) Interactive gameplay prototype.} We build \rigarena, integrating
few-step audiovisual generation with closed-loop interaction to support
continued play under persistent game rules.

\afterpage{\afterpage{\begin{figure}[t]
    \centering
    \includegraphics[width=\linewidth]{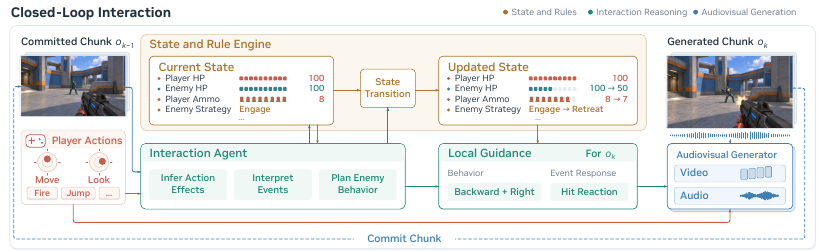}
    \caption{\textbf{Closed-loop interaction in \rig.}
    The interaction agent uses committed observations, player actions, and
    persistent state to infer events and behavior. The updated state and interaction
    proposals produce local guidance for the next audiovisual chunk, whose observations
    inform subsequent decisions.}
    \label{fig:hcworld_architecture}
\vspace{-6pt}
\end{figure}
  \afterpage{\begin{figure}[t]
    \centering
    \includegraphics[width=\linewidth]{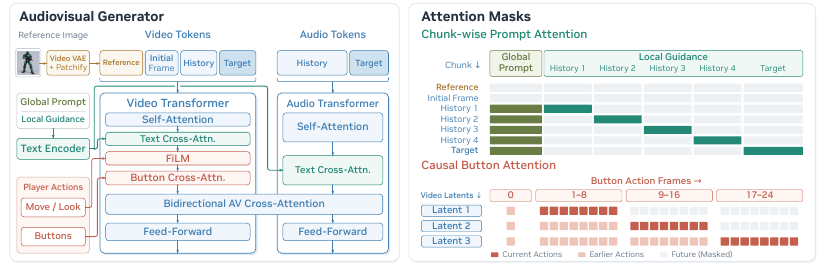}
    \caption{\textbf{Audiovisual generation with player actions and interaction guidance.}
    \emph{Left:} Global prompts and local guidance condition both Transformers,
    while player actions condition the video stream.
    \emph{Right:} Each chunk attends to the global prompt and its own local
    guidance. Each video latent aligns with eight action frames and attends
    to current and earlier actions, with future actions masked.}
    \label{fig:hcworld_generator}
\vspace{-6pt}
\end{figure}
    \afterpage{\begin{table}[!t]
\centering
\caption{\textbf{Dataset composition and annotation coverage.}
(a) Collection by route. (b) Annotation coverage across datasets.
GT denotes recorded positions and orientations. Mixed combines GT and estimated poses.
Local text describes temporally localized behavior, state labels record gameplay
variables, and controlled outcomes cover misses, nonlethal hits, and kills.}
\label{tab:dataset}
\vspace{2pt}
{\small
\setlength{\tabcolsep}{3pt}
\renewcommand{\arraystretch}{1.05}
\begin{tabular*}{\linewidth}{@{\extracolsep{\fill}}llrr@{}}
\multicolumn{4}{@{}l}{\textbf{(a) Our collection}} \\
\toprule
Route & Coverage & Recordings & Hours \\
\midrule
Basic Dynamics & Controls and local dynamics & 50,140 & 69.64 \\
Interaction & Controlled interactions & 42,000 & 47.25 \\
Interaction-Long & Actor sequences and outcomes & 1,791 & 29.87 \\
Arena & Sustained one-on-one combat & 1,800 & 30.03 \\
\bottomrule
\end{tabular*}}

\vspace{6pt}
{\fontsize{8}{9.5}\selectfont
\setlength{\tabcolsep}{2pt}
\renewcommand{\arraystretch}{1.08}
\newcommand{\dhead}[2]{\parbox[c][19pt][c]{#1}{\centering #2}}
\newcommand{\dcite}[1]{{\fontsize{7}{8}\selectfont\citep{#1}}}
\begin{tabular*}{\linewidth}{@{\extracolsep{\fill}}lcccccc@{}}
\multicolumn{7}{@{}l}{\textbf{(b) Comparison with related datasets}} \\
\toprule
Dataset & Audio & \dhead{49pt}{Action\\Labels} & \dhead{28pt}{Pose\\Labels}
& \dhead{23pt}{Local\\Text} & \dhead{25pt}{State\\Labels}
& \dhead{42pt}{Controlled\\Outcomes} \\
\midrule
CS:GO~\dcite{pearce2022counterstrike} & -- & Inputs & GT & -- & $\checkmark$ & -- \\
Gaming-500-hours~\dcite{markov2026gaming500} & -- & Inputs & -- & -- & -- & -- \\
GF-Minecraft~\dcite{yu2025gamefactory} & -- & Inputs & -- & -- & -- & -- \\
PLAICraft~\dcite{he2025plaicraft} & $\checkmark$ & Inputs & -- & -- & -- & -- \\
CrossFPS~\dcite{tong2026scope} & -- & Inputs & -- & -- & -- & -- \\
SANA-WM~\dcite{zhu2026sanawm} & -- & -- & Mixed & -- & -- & -- \\
ABot-World-0~\dcite{jiang2026abotworld0} & -- & Inputs & Mixed & -- & -- & -- \\
SolarWM~\dcite{huang2026solarwm} & -- & -- & Mixed & -- & -- & -- \\
Incantation~\dcite{zhu2026incantation} & -- & Animation IDs & -- & $\checkmark$ & -- & -- \\
WildWorld~\dcite{li2026wildworld} & -- & Animation IDs & GT & $\checkmark$ & $\checkmark$ & -- \\
\midrule
\textbf{Ours} & $\checkmark$ & Inputs & GT & $\checkmark$ & $\checkmark$ & $\checkmark$ \\
\bottomrule
\end{tabular*}}
\vspace{-6pt}
\end{table}
      \afterpage{\begin{figure}[t]
  \centering
  \includegraphics[width=\linewidth]{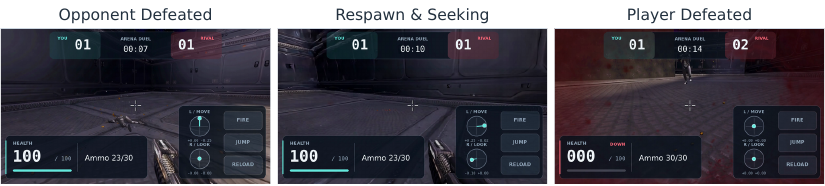}
  \caption{\textbf{A gameplay session in \rigarena.}
  Opponent defeat, respawn and seeking, and player defeat.
  The HUD is rendered from rule-engine state and overlaid on the video,
  not generated by the video model.}
  \label{fig:arena_demo}
\vspace{-6pt}
\end{figure}
\begin{table}[!t]
\centering
\caption{Model comparisons and ablations on 100 Xonotic scenarios with
50 survival and 50 death targets.
\rig (w/ Reference guidance) is a privileged offline control.}
\label{tab:interaction_results}
\fontsize{8}{9.5}\selectfont
\setlength{\tabcolsep}{2pt}
\begin{tabular*}{\linewidth}{@{\extracolsep{\fill}}lccccccccc@{}}
\toprule
\multirow{2}{*}{Method / Variant}
& \multicolumn{5}{c}{Visual Quality \& Dynamics}
& \multicolumn{4}{c}{Interaction Quality} \\
\cmidrule(lr){2-6}\cmidrule(lr){7-10}
& MS $\uparrow$ & AQ $\uparrow$ & SC $\uparrow$ & BC $\uparrow$ & DD
& EB $\uparrow$ & FSA $\uparrow$ & FSA-S $\uparrow$ & FSA-D $\uparrow$ \\
\midrule
SCOPE~{\fontsize{7}{8}\selectfont\citep{tong2026scope}} & 94.82 & 36.20 & 61.15 & 81.71 & 87.00 & 35.71 & 4.00 & 6.00 & 2.00 \\
LingBot-World 2.0~{\fontsize{7}{8}\selectfont\citep{gao2026lingbot}} & 96.50 & \textbf{53.86} & 79.99 & 92.82 & \textbf{100.00} & 55.32 & 29.00 & 58.00 & 0.00 \\
\textbf{\rig (Ours)} & \textbf{99.18} & 50.93 & \textbf{90.97} & \textbf{94.17} & 89.00 & \textbf{78.91} & \textbf{81.00} & \textbf{90.00} & \textbf{72.00} \\
\midrule
Motion guidance only & 99.17 & 50.90 & 90.85 & 94.11 & 88.00 & 76.59 & 39.00 & 78.00 & 0.00 \\
Semantic prompting & 99.13 & 50.84 & 90.92 & 94.02 & 88.00 & 75.49 & 35.00 & 70.00 & 0.00 \\
Post-hoc event updates & 99.15 & 50.67 & 90.84 & 94.01 & 88.00 & 77.74 & 64.00 & 86.00 & 42.00 \\
w/o reconciliation & 99.18 & 50.88 & 90.96 & 94.15 & 89.00 & 78.02 & 76.00 & 88.00 & 64.00 \\
\midrule
\rig (w/ Reference guidance) & 99.18 & 50.96 & 90.99 & 94.14 & 89.00 & 81.99 & 89.00 & 90.00 & 88.00 \\
\bottomrule
\end{tabular*}
\vspace{-6pt}
\end{table}
        \afterpage{\begin{figure}[!t]
  \centering
  \includegraphics[width=\linewidth]{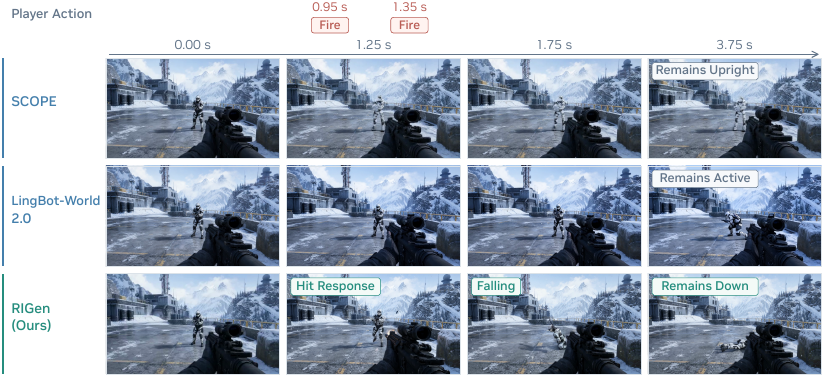}
  \caption{\textbf{Qualitative combat responses.}
  Red markers denote fire commands. Despite receiving the prescribed outcomes
  as text guidance, SCOPE and LingBot-World~2.0 keep opponents active under
  repeated fire. \rig produces a hit reaction, a fall, and a sustained
  downed state.}
  \label{fig:qualitative_comparison}
  \vspace{0pt}
  \input{figures/state_branches}
\vspace{-6pt}
\end{figure}
}}}}}}
\section{Related Work}
\label{sec:related_work}

\paragraph{Video generation and interactive worlds.}
Video and audiovisual foundation models provide appearance and dynamics
priors~\citep{kong2024hunyuanvideo,wan2025wan,ruan2023mmdiffusion,polyak2024moviegen,hacohen2026ltx2,su2026omniforcing}.
Controllable visual simulation~\citep{kim2020gamegan,menapace2021playable,menapace2022environments}
has developed into generative game environments~\citep{bruce2024genie,valevski2025gamengen,alonso2024diamond,kanervisto2025muse,decart2024oasis,guo2025mineworld}
and interactive content creation~\citep{che2024gamegenx,yu2025gamefactory,li2025gamecraft,zhang2025matrixgame,yan2025yan}.
Recent work broadens scene coverage and streaming control~\citep{he2025matrixgame2,tong2026scope,robbyant2026lingbotworld,zhu2026sanawm,huang2026solarwm},
adds multiplayer or audiovisual interaction~\citep{hu2026mira,zhang2026echowm},
and improves long-horizon consistency through memory~\citep{xiao2025worldmem,wu2026infiniteworld,sun2025worldplay,wang2026matrixgame3,alayaworld2026v11}.
We focus on persistent gameplay state and rule-governed interaction consequences.

\paragraph{Actor control and semantic interaction.}
Reactive opponents and language-guided character control extend player-driven
generation~\citep{agarwal2026combat,wang2026reactivegwm,cao2026shadowdancer,ouyang2026helloworld,tang2025gamecraft2}.
LingBot-World 2.0~\citep{gao2026lingbot} uses pilot and director agents for
character control and scene events. Incantation~\citep{zhu2026incantation}
tracks damage after generation, leaving hit/miss resolution to the video
model and risking delayed death responses during fast-paced combat.
\rig couples behavior planning with executable gameplay rules, predicts
interaction consequences before generation, and reconciles state with
committed video.

\paragraph{Explicit state and generative rendering.}
Prior systems combine structured state or game logic with generative
rendering~\citep{li2025unbounded,cheng2025animegamer,chen2025maag,li2026wildworld,lin2026stateplay,cai2026mass},
while planning-based video methods organize content before
synthesis~\citep{lin2024videodirectorgpt,lian2024lvd}.
Geometric rendering interfaces provide explicit spatial
control~\citep{huang2026alayarenderer,lin2026alayarendererflash,meng2026marionette}.
Programmable World Model~\citep{huang2026pwm} compiles state-augmented 3D
bounding boxes into rendering controls, while Code World Model~\citep{chen2026codeworld}
compiles executable state into proxy videos. \rig instead connects gameplay
state to generation through player actions and local interaction guidance,
without constructing explicit spatial trajectories.
The PixVerse Game Engine technical blog~\citep{pixverse2026gameengine}
describes a separation of game mechanics, agent orchestration, and generative
expression. We provide a concrete implementation for continuous player
control, connecting interaction prediction, executable state updates, and
video feedback, with corresponding generator training and controlled ablations.

\section{Method}
\label{sec:hcworld}

\subsection{Overview}
\label{sec:hierarchical_causality}

\rig couples three layers to align player actions, persistent state,
and generated audiovisual outcomes (Figure~\ref{fig:hcworld_architecture}).
The \emph{State and Rule Engine} maintains persistent state, applies transition
rules, and specifies strategic constraints (Section~\ref{sec:state_rules}).
The \emph{Interaction Agent} reasons about action consequences and plans actor
behavior (Section~\ref{sec:closed_loop}). The \emph{Audiovisual Generator}
models local dynamics under player controls and generation guidance
(Section~\ref{sec:scoped_generation}).

At each chunk boundary, the agent uses recent video, player controls, and
persistent state to predict interaction events and plan actor behavior.
The rule engine evaluates these events, and the updated state and planned
behavior jointly guide the next audiovisual chunk. After generation, the
agent checks which events occurred, and the rule engine uses this evidence
to determine the state for the next decision.

\subsection{Persistent State and Rule-Based Updates}
\label{sec:state_rules}

We represent gameplay as an event-driven state machine with numerical variables,
following standard state-based game AI~\citep{millington2009gameai}.
The engine maintains player and enemy health and life/death state, player
ammunition, and opponent strategy modes, together with configurable rule
parameters such as the player's per-shot damage and transition conditions.
Writing $s$ for the complete game state and $e$ for an interaction event
record, the engine maps $s$ to its successor $s'$ through
\begin{equation}
s'=\operatorname{Rules}(s,e).
\label{eq:state_transition}
\end{equation}
For example, firing requires a living player with available ammunition and
consumes a round regardless of whether the shot hits. A hit reduces the
target's health by the configured damage, for either the player or an enemy.
Reaching zero health triggers death. If an opponent survives with low health,
a configured rule can switch its strategy from engagement to retreat.
The updated strategy guides the Interaction Agent's next decision, such as
choosing a retreat direction suited to the current scene
(Section~\ref{sec:closed_loop}).

\subsection{Event Reasoning, Behavior Planning, and Visual Feedback}
\label{sec:closed_loop}

Before generating chunk $o_k$, we perform \emph{reconciliation} through
verification and state correction. The agent verifies $o_{k-1}$ using sampled
verification frames $\mathcal{I}_{k-1}$, player controls $u_{k-1}$, and local
guidance $p_{k-1}$.
The rule engine then corrects the planned state $s_k^{\mathrm{plan}}$
using feedback $f_{k-1}$:
\begin{equation}
\begin{aligned}
f_{k-1} &= \operatorname{Verify}(\mathcal{I}_{k-1},u_{k-1},p_{k-1}),\\
s_k &= \operatorname{Correct}(s_k^{\mathrm{plan}},f_{k-1}).
\end{aligned}
\label{eq:state_feedback}
\end{equation}
Using this updated state, feedback, and current player controls $u_k$,
the agent proposes events $e_k$ and behavior $b_k$:
\begin{equation}
(e_k,b_k)=\operatorname{Agent}(o_{k-1},s_k,u_k,f_{k-1}).
\label{eq:interaction_reasoning}
\end{equation}
The rule engine computes a provisional successor $s_{k+1}^{\mathrm{plan}}$
and local interaction guidance $p_k$ without overwriting $s_k$:
\begin{equation}
\begin{aligned}
s_{k+1}^{\mathrm{plan}} &= \operatorname{Rules}(s_k,e_k),\\
p_k &= \operatorname{Prompt}(s_{k+1}^{\mathrm{plan}},e_k,b_k).
\end{aligned}
\label{eq:interaction_guidance}
\end{equation}
The generator realizes this guidance as the next audiovisual chunk $o_k$
(Section~\ref{sec:scoped_generation}).
At the next boundary, verification of $o_k$ reconciles
$s_{k+1}^{\mathrm{plan}}$ with committed visual outcomes to obtain $s_{k+1}$.
These observations take precedence over conflicting predictions, and the
corrected state conditions the next decision.
Implementation scope and verification details are given in
the appendix.

\subsection{Audiovisual Generation}
\label{sec:scoped_generation}

The global prompt $g$ describes the scene and character identity shared across
chunks. Local guidance $p_k$ specifies the current chunk's actor behavior and
observable interaction outcomes, derived from the agent's proposals and
rule-based state updates (Section~\ref{sec:closed_loop}). Together, they form
text condition $c_k$. Given audiovisual context $\mathcal{H}_k$ and player
controls $u_k$, the generator produces the next chunk:
\begin{equation}
c_k=(g,p_k), \qquad
o_k\sim G_\theta(\mathcal{H}_k,u_k,c_k).
\label{eq:interaction_generation}
\end{equation}
The context $\mathcal{H}_k$ contains a first-frame anchor, a character reference,
and recent generated chunks with their controls and text conditions.
The generator denoises the target under these conditions, and the resulting
$o_k$ provides observations for the next interaction decision.

We instantiate $G_\theta$ with the LTX-2.3~\citep{hacohen2026ltx2,lightricks2026ltx23}
audiovisual backbone, using coupled video and audio transformer streams
(Figure~\ref{fig:hcworld_generator}). Continuous movement/camera controls
modulate video features through FiLM~\citep{perez2018film}, while button inputs enter through causal
cross-attention. Audiovisual attention connects the two streams and is
causal across chunks and joint within each chunk. The anchor and reference
provide visual context without text or action updates.

\paragraph{Chunk-wise prompt routing.}
In both video and audio streams, each chunk attends only to the global prompt
and its corresponding local guidance through text cross-attention
(Figure~\ref{fig:hcworld_generator}).
Historical chunks retain their original text conditions, while the target
receives the current guidance. The same routing is used in bidirectional and
causal training.

\paragraph{Training and continued generation.}
Following the bidirectional-to-autoregressive adaptation used in
CausVid~\citep{yin2025causvid}, we first train the model bidirectionally to follow player controls and
chunk-level guidance, then adapt it to autoregressive next-chunk prediction
conditioned on clean audiovisual history. Sliding-window training retains
a window of recent chunks alongside the first-frame anchor and character
reference, keeping $\mathcal{H}_k$ bounded during continued generation.
Training objectives and implementation details are provided in the appendix.

\section{Xonotic Interaction Dataset}
\label{sec:dataset}

We modify the open-source engine of Xonotic~\citep{xonotic2023} to construct
an audiovisual gameplay dataset with controlled capture and synchronized
annotations. Three complementary routes target different aspects of gameplay.
\emph{Basic Dynamics} captures how player controls affect local visual dynamics.
\emph{Interaction} covers the relationships between player actions, opponent
behavior, and combat outcomes in controlled encounters, including longer
sequences of successive encounters.
\emph{Arena} captures sustained combat between independently controlled bots
to support continued interactive play.
The collection contains 95,731 source recordings and 176.8 hours of encoded
video (Table~\ref{tab:dataset}).

Each recording provides synchronized first-person video, game audio,
player controls, and engine-grounded camera poses, together with scene
and character descriptions.
The interaction subsets pair chunk-level descriptions of opponent behavior,
visibility, and combat outcomes with engine state and event annotations.
Actor reference images are bound to the corresponding identities where used.
Table~\ref{tab:dataset} compares our dataset with existing gameplay datasets.
Our collection jointly targets local dynamics, controlled combat outcomes,
and sustained interaction. The appendix details the capture routes, alignment, filtering, and test set construction.

\section{Post-training for RIGen-Arena}
\label{sec:arena_demo}

\rigarena is an interactive one-on-one FPS prototype in which the player fights an opponent,
with either side respawning after defeat to continue the match
(Figure~\ref{fig:arena_demo}).

We initialize post-training for \rigarena from the bidirectional model trained on
the Interaction subset. We first fine-tune it bidirectionally on the Arena
subset to adapt to sustained combat. We then train the adapted model to
generate chunks autoregressively using a sliding window of recent history
together with the first-frame anchor and character reference
(Section~\ref{sec:scoped_generation}).

We adopt the two-stage distillation recipe of
Causal-rCM~\citep{zheng2026causalrcm}. Teacher-forcing consistency distillation
first produces a few-step generator. Distribution matching then trains
this generator on its own autoregressive rollouts.

We additionally train a video refiner to recover high-frequency details in
dynamic gameplay. Training pairs spatially degraded latents with the
original encodings of the same arena clips. We first train the refiner
bidirectionally with paired flow matching~\citep{lipman2023flowmatching}, then fine-tune it for causal,
chunk-wise refinement with a bounded history window.
At inference, it refines each generated chunk without access to future
chunks. We integrate the generator and refiner with the state and rule
engine and interaction agent to form \rigarena.
Training objectives, sampling schedules, and implementation details are
provided in the appendix.

\section{Experiments}
\label{sec:experiments}

This section evaluates motion adherence, combat outcomes, and visual quality
through quantitative comparisons and ablations.
\rigarena (Figure~\ref{fig:arena_demo}) and the qualitative results
(Figures~\ref{fig:qualitative_comparison} and~\ref{fig:interaction_trace})
illustrate state-dependent interactions, continued gameplay, and the world's
effects on the player.

\subsection{Quantitative Evaluation}

\paragraph{Setup.}
We compare \rig's checkpoint trained on Interaction-Long with sliding-window
autoregressive training against
SCOPE~\citep{tong2026scope} and LingBot-World~2.0~\citep{gao2026lingbot}
on 100 Xonotic combat scenarios. Test maps and enemy appearances are excluded
from all task-specific training. Enemy motion follows source annotations.
Both baselines receive prescribed combat outcomes as text guidance.
InternVL3-8B~\citep{zhu2025internvl3} is our Interaction Agent for event
reasoning and visual state correction.

\paragraph{Ablation Variants.}
\emph{Motion guidance only} retains player inputs and source motion,
without interaction guidance or an online agent.
\emph{Semantic prompting} provides state-update rules as text for the agent
to interpret and iteratively update its textual state, without an executable
rule engine or independent verification.
\emph{Post-hoc event updates} moves interaction reasoning and rule-based
state updates from before generation to after it, so the resulting guidance
conditions the next chunk rather than the current one.
\emph{Without reconciliation} disables visual correction of planned state.
\emph{\rig (w/ Reference guidance)} freezes event guidance derived from
source hit annotations and health/damage rules, without an online agent.

\paragraph{Metrics.}
Qwen3.8-27B~\citep{qwen2026qwen38} evaluates two interaction metrics offline.
\emph{Enemy Behavior} (EB) measures adherence to prescribed enemy motion
within specified time windows. \emph{Final-state Agreement} (FSA) measures
agreement between the final visible enemy state and its survival/death target.
FSA-S is the percentage of the 50 survival-target cases in which the enemy
is visibly alive at the end. FSA-D is the percentage of the 50 death-target
cases in which the enemy is visibly dead at the end. Overall FSA averages
these two rates. Incorrect or unjudgeable outcomes count as failures, so FSA
can fall below 50\%. We also report Motion Smoothness (MS), Aesthetic Quality (AQ),
Subject Consistency (SC), Background Consistency (BC), and Dynamic Degree (DD),
following VBench~\citep{huang2024vbench}. Scores are on a 0--100 scale.
Higher is better except for DD, which measures motion prevalence rather than quality.

\paragraph{Results.}
Table~\ref{tab:interaction_results} shows that \rig achieves 81\% FSA,
compared with 29\% for LingBot-World~2.0 and 4\% for SCOPE, and higher EB
than both baselines. With the generator fixed, FSA falls to 39\% with
motion-only guidance, 35\% with semantic prompting, 64\% with post-hoc
updates, and 76\% without reconciliation. These results support predicting
interaction consequences before generation, executing state updates, and
reconciling them with visual feedback. Visual metrics remain nearly unchanged
across ablations. Lower AQ than LingBot-World~2.0 may reflect our smaller
training dataset. The appendix provides further details and analysis.

\subsection{Qualitative Evaluation}

\paragraph{Combat responses.}
Figure~\ref{fig:qualitative_comparison} compares visible combat responses
from the same initial image and attack requests. In this example, \rig
produces a hit reaction followed by a fall and a sustained downed state,
while the baseline opponents remain active under repeated fire despite
receiving text guidance specifying the correct combat outcomes.

\paragraph{State-dependent interactions.}
Figure~\ref{fig:interaction_trace} varies initial enemy health, player
ammunition, and state-dependent opponent strategy from a shared starting
scene under the same player input requests.
The low-health opponent falls, depleted ammunition suppresses the third
shot, and an injured opponent switches from hold to retreat, moving left
and crouching at cover. These examples illustrate how explicit state and
strategy affect visible gameplay.

\vspace{-4pt}
\section{Conclusion}
\label{sec:conclusion}

We presented \rig, a framework for reciprocal interaction in generative
gameplay. Player actions change persistent world state, while world rules
and actor decisions affect the player. Quantitative comparisons demonstrate
more reliable combat outcomes, and qualitative results illustrate
state-dependent interactions and autonomous enemy behavior.

\vspace{-4pt}
\paragraph{Limitations and future work.}
\rig is not yet real-time, supports only single-opponent FPS gameplay,
and relies on VLM perception and reasoning for interaction decisions and
state verification. Future work will optimize generation and agent inference,
extend to multiple opponents and broader action spaces and game genres,
and improve visual grounding and reasoning reliability.

\clearpage
\subsection*{AI use statement}
We used generative AI tools to assist with drafting and
polishing the manuscript and with research execution,
specifically code implementation, debugging, experiment
scripting, and development of the dataset construction
pipelines. The authors reviewed and revised
the AI-assisted text and checked the AI-assisted code
and experimental outputs for correctness. The authors
take full responsibility for the final content of this work.

\subsection*{Ethics statement}
We construct our dataset through scripted and bot-controlled gameplay in
Xonotic, without collecting human-player data or conducting human-subject
experiments. The official Xonotic distribution is licensed under
GPL-3.0-or-later, with its DarkPlaces engine under
GPL-2.0-or-later.\footnote{\url{https://github.com/xonotic/xonotic/blob/master/COPYING}}
Code and assets remain subject to their applicable upstream licenses.
Our release will document the licenses for the dataset and accompanying
code and preserve required attribution and notices. We follow the applicable
usage terms of pretrained models. The proposed method is intended for
research purposes. Deployment should account for potential misuse of
generated content, including misleading or deceptive media.

\subsection*{Reproducibility statement}
The main text describes the proposed method and
experimental setup. Additional implementation details,
hyperparameter settings, and evaluation protocols
are provided in the appendix to support reproducibility.
The datasets and baseline methods used in our experiments
are identified and referenced in the paper.

\ificlrfinal
\section*{Acknowledgments}
This research was supported by the Theme-based Research Scheme (T45-205/21-N)
from the Hong Kong Research Grants Council (RGC), and by the Generative AI
Research and Development Centre from InnoHK. We also thank Vivix for providing
computational resources and research support during the internship. The work by
Mengfei Li, Xingtong Ge, Lunjie Zhu, and Fangyu Lin was completed during their
internships at Vivix Group Limited.
\fi

\bibliography{iclr2027_conference}
\bibliographystyle{iclr2027_conference}

\clearpage
\appendix
\appendixcontentstrue
\begingroup
\renewcommand{\contentsname}{Appendix Contents}
\setcounter{tocdepth}{2}
\tableofcontents
\endgroup
\clearpage
\section{Implementation Details}
\label{app:implementation}

\subsection{Audiovisual Generator and Action Conditioning}
\label{app:generator_details}

We build on the LTX~\citep{hacohen2026ltx2} audiovisual architecture, using
the LTX-2.3 22B dev backbone~\citep{lightricks2026ltx23}. Separate video and audio
autoencoders produce latent sequences processed by coupled transformer
streams. Each stream uses self-attention and text cross-attention, with
cross-modal attention connecting video and audio. At 20 video frames per
second, each generated chunk contains two video latent steps, corresponding
to 16 new frames, and 20 audio latent steps, corresponding to 0.8 seconds.
Generation is causal between chunks. Tokens within a chunk are denoised
jointly.

The action interface provides four continuous movement/camera channels
and supports fire, jump, and reload inputs. For each generated video latent step, the eight aligned raw-frame vectors
are concatenated into a 32-dimensional continuous-action vector $J_t$ and broadcast
over spatial locations. After projection, per-layer networks produce feature
scale and shift through FiLM~\citep{perez2018film}
\begin{equation}
h'_{\ell,t} =
\bigl(1+\gamma_\ell(J_t)\bigr)\odot h_{\ell,t}+\beta_\ell(J_t).
\label{eq:continuous_action}
\end{equation}
Buttons remain at the raw-frame rate and enter through an additional
cross-attention residual. Each generated video latent attends to its eight
aligned action frames and earlier retained actions, with future actions masked. Temporal rotary embeddings~\citep{su2024roformer}
align these queries and keys. Continuous
scale/shift outputs and button-attention output projections are initialized
to zero. The action branches act on the video stream, while cross-modal
attention connects the resulting dynamics to audio.

Writing $z_j$ for the audiovisual latents of generated chunk $j$, the causal
model uses a compact context
\begin{equation}
[\text{first-frame anchor}\mid z_{k-4:k-1}\mid T_k\mid r].
\label{eq:sliding_layout}
\end{equation}
Here $r$ denotes the encoded character reference. At steady state, four
previous chunks are clean context and only the target
chunk $T_k$ is denoised. Tokens in each generated chunk attend to tokens in
that chunk and earlier chunks.
Reference-image tokens attend only to one another and are readable directly by
the target video chunk. Direct cross-modal attention excludes the reference.
The anchor and reference receive no action or text updates. As the window
advances, old history is removed and temporal positions are rebased locally.
The same layout supports zero to three history chunks during startup.
An optional cache reuses clean-context keys and values within a chunk's
denoising steps. It is rebuilt when the window advances.

\subsection{Chunk-wise Prompt Routing and Guidance Construction}
\label{app:prompt_details}

The generator's prompt does not encode numerical health, hit counters, or internal
generation identifiers. High-frequency player controls remain in the action
tensors, while text specifies the actor behavior and consequences those
controls should realize.

Each chunk has its own text condition $c_j=(g,p_j)$. The global description
covers the environment, weapon, and character appearance. Local guidance
describes the current actor behavior, visibility, and observable interaction
outcomes. Each chunk's combined text condition is encoded independently,
and the resulting context tokens are concatenated. The text attention mask routes both video
and audio queries to the corresponding prompt, with padding
excluded and anchor/reference queries receiving no text-attention update
in the causal model. In bidirectional adaptation,
the initial frame instead shares the first generated chunk's prompt.
Historical chunks use their corresponding prompts, while the target
directly attends only to its current prompt. Historical context remains
available through audiovisual attention, which is causal across chunks
during autoregressive training and generation.

At inference, a deterministic text renderer converts the controller's
structured predictions and rule-updated state into local guidance.

\subsection{Training Objectives and Distillation}
\label{app:training_details}
\label{sec:data_training}

Generator training optimizes the audiovisual backbone and action branches
while freezing the text encoder and autoencoders.
The Interaction Agent uses InternVL3-8B~\citep{zhu2025internvl3} without
task-specific fine-tuning, while the rule engine and prompt compiler are
programmed. The loader aligns video, audio, player controls,
reference images, and chunk text conditions at shared temporal boundaries.

\paragraph{Training hyperparameters.}
We train on 16 NVIDIA GPUs with a global batch size of 16 and
BF16 mixed precision. Table~\ref{tab:training_hyperparameters} lists the
generator learning rates. Consistency distillation and distribution matching
use AdamW with $\beta=(0,0.999)$, weight decay $0.01$, and 100 warmup steps.
The learned score model uses a learning rate of $10^{-5}$.
Both refiner stages use 8-bit AdamW with a learning rate of $10^{-5}$,
$\beta=(0.9,0.999)$, weight decay $0.01$, and 100 warmup steps.
The domain LoRA has rank 384 and scaling parameter $\alpha=384$.

\begin{table}[htbp]
\centering
\caption{Generator training learning rates.}
\label{tab:training_hyperparameters}
\begin{tabular}{lcc}
\toprule
Training stage & Backbone & Action branches \\
\midrule
Basic Dynamics adaptation & $10^{-6}$ & $10^{-5}$ \\
Interaction adaptation & $5\times10^{-7}$ & $10^{-7}$ \\
Arena bidirectional adaptation & $5\times10^{-7}$ & $10^{-7}$ \\
Sliding-window autoregressive training & $10^{-6}$ & $10^{-5}$ \\
Consistency distillation & $10^{-6}$ & $10^{-6}$ \\
Distribution matching & $10^{-5}$ & $10^{-5}$ \\
\bottomrule
\end{tabular}
\end{table}

\paragraph{Action and prompt adaptation.}
Basic Dynamics data teach control-conditioned local dynamics, while the
Interaction data add chunk-local behavior and outcome guidance. Arena
post-training starts from the Interaction-trained bidirectional model and
adapts it to sustained combat before causal training and distillation.
Bidirectional adaptation uses chunk-specific text conditions and a linear-path
flow-matching objective~\citep{lipman2023flowmatching,liu2023rectifiedflow}.
For modality $m\in\{v,a\}$, a clean latent
sequence $z^m$ is perturbed as
$z^m_\sigma=(1-\sigma)z^m+\sigma\epsilon^m$, with Gaussian noise
$\epsilon^m$. The model predicts the flow velocity $\epsilon^m-z^m$
and is trained by minimizing
\begin{equation}
\mathcal{L}_{\mathrm{bi}} =
\sum_{m\in\{v,a\}}\lambda_m
\mathbb{E}\left[
\left\|v_\theta^m(z_\sigma,\sigma;u,c)
-(\epsilon^m-z^m)\right\|_{\Omega_m}^{2}
\right],
\label{eq:bidir_training}
\end{equation}
where $c$ denotes the sequence of chunk text conditions and the norm averages
supervised tokens, excluding the conditioning video
frame and ineligible audio samples. The text attention mask restricts each
generated chunk to its own text condition, while audiovisual self- and cross-attention remain
bidirectional at this stage.

\paragraph{Sliding-window autoregressive training.}
We initialize the causal model from the adapted checkpoint and train the
layout in Equation~\ref{eq:sliding_layout}. The anchor, reference, and available
history chunks remain clean. Noise is applied only to the next chunk.
For clean target latents $z_k$, Gaussian noise $\epsilon_k$, and noise
level $\sigma\in[0,1]$, the noisy input and predicted clean estimate are
\begin{equation}
\begin{aligned}
z_{k,\sigma}&=(1-\sigma)z_k+\sigma\epsilon_k,\\
\widehat{z}_k&=z_{k,\sigma}
-\sigma v_\theta(z_{k,\sigma},\sigma\mid\mathcal{H}_k,u_k,c_k).
\end{aligned}
\label{eq:conditional_denoising}
\end{equation}
Here $v_\theta$ predicts flow velocity, and $\mathcal{H}_k$, $u_k$, and $c_k$
follow Equation~\ref{eq:interaction_generation}. Superscripts $v$ and $a$
denote video and audio. The target-chunk reconstruction objective is
\begin{equation}
\mathcal{L}_{\mathrm{AR}}=
\mathbb{E}\!\left[w(\sigma)\left(
\operatorname{MSE}(\widehat{z}_k^v,z_k^v)
+\lambda_a\operatorname{MSE}(\widehat{z}_k^a,z_k^a)
\right)\right],
\label{eq:ar_training}
\end{equation}
where $w(\sigma)=\exp[-2(\sigma-0.5)^2]$ weights the noise level and
$\lambda_a=0.16$ weights the audio loss. Squared errors are averaged over
target tokens only.
This stage uses ground-truth history.
Bidirectional and autoregressive adaptation share the chunk-wise prompt
routing used at inference, but only autoregressive adaptation enforces
temporal causality through audiovisual attention.

\paragraph{Consistency distillation and startup coverage.}
For few-step generation, the implementation follows the teacher-forcing
consistency initialization in Causal-rCM~\citep{zheng2026causalrcm}.
This recipe builds on Consistency Models~\citep{song2023consistency},
sCM~\citep{lu2025scm}, and rCM~\citep{zheng2026rcm}.
A frozen autoregressive teacher advances a noisy sample between nearby noise
levels. At the advanced point, the current student produces a clean estimate
without gradients, which serves as the consistency target for its prediction
at the starting point. Video and audio both contribute to this loss.
Checkpoint EMA is maintained separately.

Startup and continued generation use the same sliding-window context layout. The
training recipe alternates an update with four history chunks and an update
with zero, one, two, or three history chunks.

The final Arena recipe further distills text classifier-free
guidance~\citep{ho2022cfg} into the student. The teacher combines conditional and text-unconditional
predictions while retaining player actions and visual context. The student
uses a single conditional forward pass per denoising step.

\paragraph{Self-forcing distribution matching.}
After consistency distillation, we apply self-forcing distribution
matching~\citep{huang2025selfforcing} as the second stage of the
Causal-rCM~\citep{zheng2026causalrcm} recipe. The objective follows
DMD~\citep{yin2024dmd} and DMD2~\citep{yin2024dmd2}, with bidirectional-teacher
supervision of an autoregressive student as in CausVid~\citep{yin2025causvid}.
One student starts from
the initial frame, generates four startup chunks, and continues for five
additional chunks using the same reference, chunk-wise prompt routing, and sliding
context. No ground-truth history is inserted into this rollout. Both the
startup chunks and the later chunks receive distribution-matching supervision.

A frozen bidirectional score model and a learned score model for student
samples evaluate the same perturbed rollout with aligned controls and
semantic conditions. These score models receive no reference-image tokens,
so their text conditions describe character appearance explicitly.
The final Arena recipe applies text guidance to the
frozen score model, while the learned score model and student remain
conditional. Their normalized difference in denoised predictions
defines a stop-gradient target. The generator then replays the recorded final
denoising step of each chunk to optimize a masked video--audio regression
loss against that target. Rollout generation runs without gradients, avoiding storage of its full
computation graph.

\paragraph{Video refiner training.}
We construct temporally aligned training pairs from the same arena clip.
Its original VAE encoding serves as the target. To construct the degraded
input, we spatially downsample the RGB frames, encode them, and apply a
frozen LTX-2.3 $2\times$ spatial upscaler v1.1~\citep{lightricks2026ltx23}
to restore the target spatial size.
We train a domain LoRA~\citep{hu2022lora} on the refiner with truncated
flow matching~\citep{lipman2023flowmatching} between
the original target and a noise-perturbed upsampled latent.
The refiner is first trained bidirectionally. Causal adaptation starts from
this checkpoint, conditions on the first frame, character reference, and
up to four clean history chunks, and applies loss only to the current
chunk. Training uses ground-truth history, while deployment uses previously
refined outputs. Both stages optimize only the domain LoRA, keeping the upsampler and
backbone frozen. The official LTX-2.3 distilled LoRA
v1.1~\citep{lightricks2026ltx23} is disabled for this training loss and enabled
for few-step refinement at inference.

\paragraph{Inference settings.}
The distilled Arena generator uses four denoising steps with noise levels
$(1,0.9375,0.8333,0.625,0)$.
Refinement uses two denoising steps with noise levels
$(0.725,0.421875,0)$, a classifier-free guidance scale of 1, and no
spatiotemporal guidance.

\subsection{State Updates and Visual Verification}
\label{app:verification_details}

\paragraph{Prediction and state commitment.}
At a chunk boundary, the controller retains the corrected state and uses
predicted events to compute provisional outcomes and local guidance.
Generation does not commit those predictions to the persistent ledger.
After decoding, verification produces the feedback used to obtain the next
retained state in Equation~\ref{eq:state_feedback}. The implementation
recomputes this state from the saved pre-generation state and observed
events, preventing predicted and verified damage from being counted twice.
The first chunk uses the initial frame and configured state without a
preceding verification step.
The retained gameplay state is stored independently of the generator's
sliding visual context. When earlier frames leave that context, their
recorded interaction consequences remain available to subsequent rule
updates and enter generation through the current guidance.

\paragraph{Observation selection and shot prediction.}
The online Arena implementation uses decoded RGB frames directly, without
an observation refiner. Within each chunk, it selects the first frame of
each uninterrupted firing interval and every fourth frame thereafter
(5 Hz at 20 FPS). Before generation, the latest available frame is the
preceding chunk's last frame, or the initial image at startup. For each
selected firing time, a calibrated control-to-pixel mapping estimates the
image displacement from the intervening controls. We translate the latest
frame accordingly, take a $320\times320$ crop centered on the screen, and
overlay a hollow green aiming ring. The Interaction Agent receives this
projected crop and a separate enemy appearance reference, and predicts
whether the enemy's body overlaps the marked aiming region.

\paragraph{Visual verification.}
After generation, we decode the corresponding actual firing frames,
the chunk's final frame, and the frame three positions before it.
Together with the pre-generation frame, these form the sampled observations
$\mathcal{I}_{k-1}$. A firing-frame audit receives the selected firing
frames in temporal order, with the same ring overlay, alongside the
pre-generation and two late frames. Unlike prediction, verification uses
full frames rather than center crops. The ring radius is 5\% of the shorter
image dimension and leaves its interior unobstructed.
A separate state query places the final frame first and uses the
pre-generation and earlier late frame only as temporal context. It
classifies the enemy as alive, dead, or not visible in the final frame.
Both queries receive an appearance reference for identity only. Full
frames are padded to a $2{:}1$ canvas and the reference to a square,
without cropping their contents. Aim markers appear only in the VLM
inputs. The requested outcome is the condition being checked, not visual
evidence, and verification uses separate calls from prediction.

The rule engine specifies intended transitions, while committed visual
outcomes ground the state used for subsequent interaction. When an observed
outcome contradicts a prediction, the controller revises the affected state
and derives subsequent guidance from that revision. Numerical attributes
remain rule-maintained rather than directly estimated from images, and
occlusion alone is not treated as death.

\paragraph{Action constraints and opponent behavior.}
Ammunition gates the executed fire commands and is consumed when a shot is
permitted, independently of whether it hits. Before an enemy can attack,
the Interaction Agent checks visual preconditions, including whether the
enemy is visible and facing the player. When these conditions hold, attacks
are sampled at a configured rate, and each attack hits with a preset
probability. Successful hits subtract a fixed
per-shot damage from player health. Reaching zero health disables player
controls and adds the corresponding terminal outcome to guidance.
For opponent behavior, the agent assesses movement feasibility from the
current scene. The behavior policy combines this assessment with retained
state and the configured strategy to select a movement description.
Attack rules and strategy-transition conditions can be edited in the rule layer.
Numerical state remains in the controller, while the generator receives
observable behavior and outcome descriptions. World-side events thus
combine visual assessment of the current scene with executable game rules
that update player state and constrain subsequent controls.

\subsection{Design Rationale}
\label{app:design_rationale}

Reciprocal interaction concerns the causal effects of player actions on the
world and of world events on the player. It does not require every game
mechanic to be learned. Executable rules make damage, ammunition, and
termination conditions explicit and configurable, while scene-conditioned
behavior reasoning determines how actors act under those conditions.
Their combination connects world-side decisions to persistent consequences
for the player during continuous interaction.

Realizing these interactions also requires a generator that can express
their consequences. Explicit state specifies intended outcomes, and text
guidance communicates them, but the generator must learn to realize them
as visible behavior. We therefore develop interaction data, train the
conditioned generator, and connect event reasoning, executable updates,
and visual feedback in an online loop. The contribution is the reciprocal
generative gameplay enabled by this coordinated system, including its data
and training procedure. Reconciliation supplies feedback within this loop
by grounding subsequent decisions in committed observations.

\section{Dataset Construction}
\label{app:dataset}

All routes share the capture and alignment pipeline below. Counts refer to
accepted source recordings, excluding candidate jobs, prompt variants, and
overlapping training windows.

\subsection{Shared Capture and Alignment}

\paragraph{Execution and recording.}
We specify the map, assets, initial conditions, actor policies, and player
controls before capture. The engine records gameplay demos together with
applied inputs, frame timing, state, and interaction events.
Requested controls and realized events are retained separately, with damage
labels taken from realized engine events. Offline replay renders the recorded
trajectory and game audio under a fixed client configuration. Presentation
settings, including HUD, crosshair, and effect suppression, are versioned
per route.
Renderer telemetry records camera position and orientation, transforms,
and projection parameters, synchronized with simulation ticks.

\paragraph{Time and action interface.}
The capture pipelines use a 64\,Hz simulation/control clock. Warm-up is
removed using the recorded capture boundary before aligning model time
zero. The model-facing stream uses 20 video frames per second, with four
continuous movement/camera channels and fire, jump, and reload inputs
under the shared action contract. Camera and movement values follow the recorded
applied controls. Button resampling preserves short active runs that might
otherwise disappear between sampled frames. In the long Interaction release,
the action profiles explicitly remove 64 warm-up ticks and align row zero
with the first video frame.

Native media differ across routes. Basic Dynamics stores 320 frames at 64\,fps
and $1280\times720$ resolution. Short and long Interaction sequences store 81 and 1,201
frames at 20\,fps and $1280\times720$. Arena stores 1,201 frames at 20\,fps
and $1664\times960$. Representative files in all four subsets contain
48\,kHz stereo game audio. The model loader resizes video and resamples audio
according to the training configuration. The generator training configuration uses
$832\times480$ video and 16\,kHz stereo audio.
The 81-frame layout comprises an initial frame and five 16-frame chunks.
The 1,201-frame layout similarly supports 75 chunks.

\paragraph{Semantic labels and packaging.}
We use Gemini 3.1 Flash-Lite~\citep{google2026gemini31flashlite}
(\texttt{gemini-3.1-flash-lite}) to separately annotate map style, weapon
appearance and operation, and player and enemy appearance.
Interaction releases add chunk-local motion,
visibility, damage, and death descriptions derived from telemetry.
Prompts distinguish no damage, nonlethal damage, a new death event, and an
already-dead actor. Off-screen intervals are described using visibility
records rather than invented visible motion. Player controls remain in
action tensors. Enemy movement and jumping belong to semantic conditions.
Where reference images are used, the manifest binds them to the appropriate
actor and time interval.

\subsection{Basic Dynamics}

Basic Dynamics contains 50,140 five-second clips across 28 maps and nine
weapons, including 39,539 without an enemy and 10,601 with a non-damaging
enemy distractor.
Movement, camera changes, firing, and local collision responses provide
the principal supervision. The supplement emphasizes pure idle,
idle-to-action, action-to-idle, interrupted action, smooth routes, and
wall impacts.

Route validation and camera filtering remove unsuitable trajectories.

\subsection{Interaction}

\paragraph{Short interactions.}
The short Interaction subset contains 42,000 recordings across five maps,
nine weapons, and 42 model--skin--color appearances.
Capture varies distance, opponent movement,
direction changes, stopping, jumping, visibility transitions, and firing.
Firing times are sampled deterministically from the scenario seed.
Weapon-specific schedules include isolated shots, repeated taps, and
bounded bursts. Opponent motion and jump programs are specified separately
from player controls.

Outcome labels follow the realized engine events, yielding 6,535 no-fire examples,
16,666 misses, 14,496 nonlethal-hit examples, and 4,303 kills.
The visibility allocation is 37,800 continuously visible examples,
2,100 exit-and-hold examples, and 2,100 exit-and-return examples.
The annotation identifies visibility transitions within each
0.8-second chunk. The underlying timestamped records also support
constructing annotations over windows of arbitrary duration within each
recording. This release describes view-frustum transitions, not
geometric occlusion behind arbitrary cover. Crosshair and casing effects are
disabled in its capture configuration.

Each recording has one
81-frame bidirectional view and one five-chunk autoregressive view.

\paragraph{Long interaction sequences.}
Interaction-Long contains 1,791 accepted recordings with blood-hit effects
disabled during capture.
Each episode contains four actor slots, arranged either as distinct actors
or a controlled-return sequence, and zero, one, or two scheduled kills.
The accepted collection contains 900 controlled-return and 891
distinct-actor episodes.

The maps are Bromine, Courtfun, Geoplanetary, and Stormkeep. The weapons
are MachineGun, Rifle, and Shotgun. Scripted trajectories undergo collision
and ground-support checks. Camera schedules use frozen calibration traces,
with scan and aiming phases, rather than live feedback that tracks the
current enemy.

The reference bank contains 42 appearances with five clean views each.
Bindings expose the current actor reference, retain the outgoing reference
through its despawn transition, and mask intervals without an actor.
Each source episode provides fifteen 81-frame bidirectional windows.
The autoregressive view keeps each full sequence and its 75 chunk prompts.

\subsection{Arena: Sustained Combat}

Arena contains 1,800 one-minute
bot-versus-bot recordings in a compact arena, using MachineGun combat and
a fixed opponent appearance. Independent bot controls produce ongoing
movement and exchanges instead of prescribing a short outcome sequence.
The capture configuration bounds locomotion and camera motion and removes
crosshair and pickup clutter.

The release constructs the action stream from authoritative gameplay
records. Chunk descriptions summarize opponent motion, attacks, visibility,
nonlethal damage, kills, and player death from telemetry aligned with the
rendered camera. Fully off-screen chunks suppress visible-motion claims.
The manifest provides 75 prompts per episode and a fixed opponent
reference binding.
Arena trains \rigarena.

\subsection{Quality Control and Test Set Construction}

Capture validation checks route geometry, camera behavior, and consistency
between scheduled interactions and recorded events. Manifest validation
checks media metadata, finite and aligned controls, complete prompt
schedules, and reference bindings.

\paragraph{Test set construction.}
We reserve 100 Interaction recordings for quantitative evaluation. Their
maps and enemy appearances are excluded from all task-specific training
subsets, including Basic Dynamics, Interaction, Interaction-Long, and Arena,
and are not used in subsequent adaptation or distillation.

\section{Evaluation Details}
\label{app:evaluation}

\subsection{Evaluation Protocol}

The 100 Xonotic scenarios comprise 50 survival targets and 50 death targets.
All \rig variants in Table~\ref{tab:interaction_results} use the same
Interaction-trained generator after sliding-window autoregressive training
on Interaction-Long, without Arena post-training or distillation.
SCOPE and LingBot-World~2.0 receive the prescribed
combat outcomes directly as text guidance.

Enemy motion is supplied by source annotations in place of online behavior
selection. The interaction agent handles hit, injury, and death reasoning
and correction of predicted state.
All variants use their own generated video history. Online interaction
variants compute guidance from that history, while \rig (w/ Reference guidance)
uses frozen reference guidance.

Fixing enemy motion gives methods common behavior requirements and focuses
the comparison on motion adherence and combat-outcome realization, separating
behavior selection from its execution in generated video.

We combine controlled experiments, state-dependent branching examples, and
continuous closed-loop gameplay. Controlled experiments compare the
realization of specified interaction outcomes. Branch comparisons show how
state and strategy affect generated behavior, while \rigarena demonstrates
the mutual effects of player actions and world events during continued play.
Together, these complementary evaluations examine outcome realization,
state dependence, and reciprocal interaction in the implemented system.

The external comparisons evaluate complete systems, including their data
and training recipes. Learning to realize interaction outcomes is part of
the proposed method. Specifying an outcome in text alone does not ensure
that a generator can express it. We therefore assess the resulting system
capability against external models and use fixed-generator ablations to
examine the online interaction mechanisms separately.

\paragraph{Evaluating generated trajectories.}
Once a generated trajectory diverges from its source recording, source-engine
hit and visibility labels cannot automatically serve as ground truth for the
generated scene. Rule execution can be checked against configured transitions,
while judging whether a generated attack hits requires independent evidence
from the generated video itself. Autonomous behavior also admits multiple
valid choices under the same state, calling for task-based criteria that
accommodate alternative behaviors. Our benchmark uses prescribed motion and outcome
targets with video-based assessment. Event-level visual correctness and
autonomous decision quality require annotations or task criteria for the
generated trajectories, rather than direct reuse of source-engine labels.

\subsection{Metrics}

Qwen3.8-27B~\citep{qwen2026qwen38} judges complete generated videos offline.

\paragraph{Enemy Behavior.}
Enemy Behavior averages
satisfaction of directional motion checks within predefined time windows,
first within each video and then equally across the 100 scenarios. A check
receives credit only when the judge reports it as satisfied and cites visual
evidence within its time window. Lateral and depth motion are checked
separately, relative to the environment rather than camera-induced image
motion. For death-target cases, the prescribed lethal chunk and subsequent motion
requirements are excluded for all methods using the same fixed windows.

\paragraph{Final-state Agreement.}
Final-state Agreement compares the enemy's visible state in the final
0.8 seconds with its preassigned survival or death target. The target is
withheld from the judge. Unreadable outcomes and failures remain in the
denominator and receive no agreement credit.
Overall FSA uses all 100 cases. Survival-target and death-target success
rates each use the corresponding 50 cases, so overall FSA is their mean.
FSA tests whether the prescribed final combat outcome is actually realized.
Locally plausible motion and high visual quality do not establish that an
opponent survives or dies as required, making outcome agreement a distinct
interaction criterion. EB measures motion adherence, while qualitative
sequences illustrate intermediate combat responses.

\paragraph{Visual quality and dynamics.}
The five visual metrics use the official dimension-specific VBench
pipelines on complete generated videos. Each reported score is the mean
per-video normalized score multiplied by 100. Dynamic Degree is the
percentage of videos meeting the official motion threshold.

The system also generates audio jointly with video.

\subsection{Ablations and Reference Comparison}

The ablation rows in Table~\ref{tab:interaction_results} compare inference-time variants using the
same video checkpoint, initial conditions, player inputs, and sampling settings.
\begin{enumerate}
    \item \textbf{Full \rig.} Predict interaction consequences before
    generation, apply executable rules, and guide the current chunk.
    Observe the generated chunk and reconcile predicted state with its
    visible outcome before the next decision.
    \item \textbf{Motion guidance only.} Retain player inputs, scene,
    appearance, and source motion annotations. Remove hit, injury, and death
    guidance and the online interaction VLM.
    \item \textbf{Semantic prompting.} The VLM maintains persistent textual
    state and selects the next chunk's combat outcome. Rules are supplied
    as text, without an executable rule engine or an independent
    verification-and-correction module. This variant compares agent-managed
    state updates with executable rules and separate visual reconciliation.
    \item \textbf{Post-hoc event updates.} Generate first, observe events,
    apply rules, and pass consequence guidance to the following chunk.
    There is no pre-generation prediction of current-chunk consequences.
    \item \textbf{Without reconciliation.} Retain pre-generation prediction
    and executable state updates, but do not correct the planned state from
    post-generation verification. The next reasoning step still sees the
    newly generated frames.
    \item \textbf{\rig (w/ Reference guidance).} Freeze reference event
    guidance in advance and remove online VLM calls. Death instructions are
    derived from source hit annotations combined with configured health and
    damage rules. They specify a counterfactual task answer, not ground-truth
    death in the source recording. The guidance is not revised from generated
    observations, but autoregressive video context uses this variant's own
    outputs.
\end{enumerate}

\subsection{Runtime}

\begin{table}[!ht]
\centering
\begin{minipage}{0.64\linewidth}
\centering
\caption{Generation throughput at $832\times480$.
Parameters count the generation backbone.
Both \rig settings include refinement.}
\label{tab:runtime_results}
\small
\setlength{\tabcolsep}{6pt}
\begin{tabular}{@{}lrr@{}}
\toprule
Setting & Parameters & FPS \\
\midrule
SCOPE & $\sim$5B & 4.12 \\
LingBot-World 2.0 & 14B & 4.47 \\
\rig (Replay) & 22B & 4.41 \\
\rig (Online) & 22B & 3.93 \\
\bottomrule
\end{tabular}
\end{minipage}
\end{table}

Runtime is measured on a single NVIDIA GPU.
In Table~\ref{tab:runtime_results}, Replay disables online VLM calls and
reuses the same guidance, while Online runs the interaction loop. Both
include refinement and remain below the target playback rate of 20 FPS.

\end{document}